\documentclass[11pt,a4paper]{article}
\usepackage{acl2017}
\usepackage{times}
\usepackage{latexsym}
\usepackage[english]{babel}
\usepackage{blindtext}
\usepackage{adjustbox}

\usepackage{url}

\newcommand{\bleu}{\textsc{\small bleu}}
\newcommand{\rouge}{\textsc{\small rouge}}

\newcommand{\ter}{\textsc{\small ter}}

\aclfinalcopy 

\usepackage[normalem]{ulem}
\definecolor{darkgreen}{rgb}{0.0, 0.5, 0.0}

\makeatletter
\def\ODhl#1{\bgroup \markoverwith{\lower3.5\p@\hbox{\sixly \textcolor{darkgreen}{\char58}}}\ULon{#1}}
\font\sixly=lasyb10 scaled 652 
\makeatother
\def\ODdel#1{\bgroup\markoverwith{\textcolor{darkgreen}{\rule[0.5ex]{2pt}{1pt}}}\ULon{#1}}

\title{Extended Abstract for WiNLP}
\title{Addressing Bottlenecks for Data-driven Natural Language Generation}
\title{Future Directions for Data-driven Natural Language Generation}
\title{Data-driven Natural Language Generation: Paving the Road to Success}

\author{Jekaterina Novikova, Ond\v{r}ej Du\v{s}ek \and Verena Rieser\\
Interaction Lab \\
Heriot-Watt University \\
Edinburgh, EH14 4AS, UK \\
  {\tt \{j.novikova, o.dusek, v.t.rieser\}@hw.ac.uk}}

\date{}

\begin{document}
\maketitle
\begin{abstract}

We argue that there are currently two major bottlenecks 
 to the commercial use of statistical machine learning approaches for natural language generation (NLG): (a) The lack of reliable automatic evaluation metrics for NLG, and (b) The scarcity of high quality in-domain corpora. We address the first problem by thoroughly analysing current evaluation metrics and motivating the need for a new, more reliable metric. 
 The second problem is addressed  by presenting a novel framework for developing and evaluating a high quality corpus for NLG training. 
\end{abstract}

\section{Evaluation metrics for NLG}
\label{sec:metrics}

Up to 60\% of  NLG research published between 2012--2015 relies on automatic evaluation measures, such as \bleu{} \cite{gkatzia:enlg2015}. The use of such metrics is, however, only sensible if they are known to be sufficiently correlated with human preferences, which is not the case, as we show 
  in the most complete study to date, across metrics, systems, datasets and domains. 

We evaluate three end-to-end NLG systems: RNNLG \cite{wen:emnlp2015}, TGen \cite{jurcicek:2015:ACL} and LOLS \cite{vlachos:coling2016}, using a large number of 21 automated metrics. 
The metrics are divided into groups of word-based metrics (WBMs, such as {\bf \ter} \cite{ter},
{\bf \bleu} \cite{papineni2002bleu},
{\bf \rouge} \cite{lin2004rouge}, semantic similarity \cite{han2013umbc} etc.) and grammar-based metrics (GBMs, such as readability, characters per utterance and per word, syllables per sentence and per word, number of misspellings etc.). To assess the metrics' reliability, we calculate the 
Spearman correlation between the metrics and human ratings for the same natural language (NL) utterances, the accuracy of relative rankings and conduct a detailed error analysis. 

\begin{table}[tp]
\begin{center}
\begin{adjustbox}{max width=0.49\textwidth}
\begin{tabular}{l|ll|ll}
 & \multicolumn{2}{l|}{Lexical richness} & \multicolumn{2}{l}{Syntactic 
 complexity} \\
Dataset & LS & MSTTR & Level 0-1 & Level 6-7 \\
\hline \hline
our corpus & \textbf{0.57} & \textbf{0.75} & \textbf{46\%} & \textbf{16\%} \\
SFRest & 0.43 & 0.62 & 47\% & 13\% \\
SFHot & 0.43 & 0.59 & 51\% & 15\% \\
Bagel & 0.42 & 0.41 & 50\% & \textbf{16\%}
\end{tabular}
\end{adjustbox}
\end{center}
\caption{Lexical richness and syntactic variation for the collected corpus and other popular datasets. 
\textit{LS} measures the proportion of less frequent words in the text, \textit{MSTTR} measures the type-token ratio normalised by the size of the corpus. For D-level complexity, \textit{Level 0-1} include syntactically simple sentences, \textit{Level 6-7} include the most complicated sentences.}
\label{tab:res}
\end{table}

The results reveal that no metric produces an even moderate correlation with human ratings (max.\ 
$\rho=0.33$), independently of dataset, system, or aspect of human rating. WBMs make two strong assumptions: They treat human-generated NL references as a gold standard which is correct and complete. We argue that these assumptions are very often invalid for corpus-based NLG, especially when using crowdsourced datasets. Grammar-based metrics, on the other hand, do not rely on human-generated references and are not influenced by their quality. However, these metrics can be easily manipulated with grammatically correct and easily readable output that is unrelated
to the input. 
Our study clearly demonstrates the need for
more advanced metrics, as used in related fields, e.g.\ MT \cite{specia:MT2010}. 


\section{Corpus development and evaluation}
\label{sec:corpus}

Recent advances in corpus-based NLG \cite{jurcicek:2015:ACL,wen:emnlp2015,Mei:NAACL2015,Wen:NAACL16,Dusek:ACL16,vlachos:coling2016} require costly training data, consisting of meaning representations (MRs) paired with corresponding NL texts. In our work, we propose a novel framework for crowdsourcing high quality NLG training data, using automatic quality control measures and evaluating two 
types of MRs, pictorial and textual, used to elicit data \cite{novikova2016crowd}. 

When collecting corpora for training NLG systems, especially when using crowd workers, the following challenges arise:

\begin{itemize}
\item[(1)] How to ensure 
the required high quality of the collected data?
\vspace{-0.2cm}
\item[(2)] What types of meaning representations can elicit spontaneous, natural and varied data from crowd workers?
\end{itemize}

To address (1), we filter the crowdsourced data using a combination of automatic and semi-manual validation procedures, as described in \cite{novikova2016crowd}.
We validate the data by selecting native English participants, allowing only well formed English sentences to be submitted, and measuring the semantic similarity of a collected NL utterance and an associated MR. 
%
Using this framework, we collected a dataset of 50k instances in the restaurant domain, which is 10 times bigger than datasets currently used for NLG training, e.g.\ SFRest and SFHot \cite{wen:emnlp2015} 
 or Bagel \cite{mairesse:acl2010}. 

To evaluate the quality of the collected corpus, we analyse the data with regards to lexical richness and syntactic variation and compare our results to other popular datasets in similar domains, i.e.\ SFRest, SFHot and BAGEL.
We use the Lexical Complexity Analyser 
 \cite{lu2012relationship} to measure various dimensions of lexical richness and variation, such as \textit{mean segmental type-token ratio} (MSTTR) and \textit{lexical sophistication} (LS). The results of lexical analysis (see Table~\ref{tab:res}) show that our  corpus is lexically more diverse and as such, considerably more complex. 
In order to evaluate syntactic variation and complexity of NL references in our corpus, we use the D-Level Analyser \cite{lu2009automatic}. 
Table~\ref{tab:res} shows that our collected corpus has the highest proportion of  complex sentences (54\% of sentences scored above level 1). 
 At the same time, the proportion of  syntactically complex sentences (levels 6 and 7) is one of the highest. 

Furthermore, our dataset requires content selection in 40\% of the cases. In contrast to the other datasets, crowd workers were asked to verbalise all the \textit{useful} information from the MR and were allowed to skip an attribute value  considered unimportant. 
 As such, learning from this dataset promises more natural, varied and less template-like system utterances.

To address challenge (2) in corpus development, we conduct a principled study regarding the trade-off between semantic expressiveness of the MR and the quality of crowdsourced utterances elicited. 
In particular, we investigate translating textual MRs (presented in the form of logic-based dialogue acts, such as ``inform(name = the Wrestlers, price range = cheap, customer rating = low)'') into pictorial representations as used in, e.g. \cite{Williams:2007,black-EtAl:2011:SIGDIAL2011}. We show that pictorial MRs result in better quality NLG data than logic-based textual MRs: utterances elicited by pictorial MRs are judged as significantly more natural, more informative, and better phrased, with a significant increase in average quality ratings (around 0.5 points on a 6-point scale), compared to the logical MRs (see Table~\ref{tab:text-picMR}). Pictorial MRs also result in more spontaneous, natural and varied utterances.
 This is probably due to crowd workers not being primed by lexical tokens. Moreover, as the MR becomes more complex, the benefits of pictorial stimuli increase.

\begin{table}[tp]
\centering
\begin{adjustbox}{max width=0.49\textwidth}
\begin{tabular}{l|l|l}
                & Logic-based MR & Pictorial MR \\
                \hline \hline
informativeness & 4.28**         & \textbf{4.51}**       \\
naturalness     & 4.09**         & \textbf{4.43}**       \\
phrasing        & 4.01**         & \textbf{4.40}**      
\end{tabular}
\end{adjustbox}
\caption{Human evaluation of the data collected with logic-based and pictorial MRs (** denotes $p<$0.01)}
\label{tab:text-picMR}
\end{table}

\section{Conclusion and future work}

Our work addresses two major bottlenecks of current data-driven NLG: Reliable automatic evaluation and efficient high-quality data collection.
While our work shows that we can effectively crowdsource data of sufficient quality to train NLG algorithms -- in particular, using pictorial representations that reduce bias and elicit more syntactically varied and lexically rich data -- our work also clearly demonstrates the need for more advanced evaluation metrics. We see our work as a first step towards reference-less evaluation for NLG by introducing grammar-based metrics.

\section*{Acknowledgements}
\vspace{-0.2cm}
This research received funding from the EPSRC projects  DILiGENt (EP/M005429/1) and  MaDrIgAL (EP/N017536/1). The Titan Xp used for this research was donated by the NVIDIA Corporation.
\bibliography{acl2017}
\bibliographystyle{acl_natbib}

\end{document}